\documentclass[letterpaper, 10 pt, journal, twoside]{IEEEtran}
\usepackage{amsmath,amsfonts}
\DeclareMathOperator*{\argmax}{arg\,max}
\usepackage{amssymb} 
\usepackage{algorithm}
\usepackage{algorithmic}
\usepackage{booktabs}
\usepackage{cite}
\usepackage{epsfig} 
\usepackage{graphics} 
\usepackage{hyperref}
\usepackage{array}
\usepackage[caption=false,font=normalsize,labelfont=sf,textfont=sf]{subfig}
\usepackage{textcomp}
\usepackage{stfloats}
\usepackage{url}
\usepackage{verbatim}
\usepackage{graphicx}
\def\BibTeX{{\rm B\kern-.05em{\sc i\kern-.025em b}\kern-.08em
    T\kern-.1667em\lower.7ex\hbox{E}\kern-.125emX}}
\usepackage{balance}
\usepackage{tikz}
\usepackage{tikz,xcolor,hyperref}

\definecolor{lime}{HTML}{A6CE39}
\DeclareRobustCommand{\orcidicon}{%
	\begin{tikzpicture}
	\draw[lime, fill=lime] (0,0) 
	circle [radius=0.16] 
	node[white] {{\fontfamily{qag}\selectfont \tiny ID}};
	\draw[white, fill=white] (-0.0625,0.095) 
	circle [radius=0.007];
	\end{tikzpicture}
	\hspace{-2mm}
}

\foreach \x in {A, ..., Z}{%
	\expandafter\xdef\csname orcid\x\endcsname{\noexpand\href{https://orcid.org/\csname orcidauthor\x\endcsname}{\noexpand\orcidicon}}
}

\newcommand\copyrighttext{%
  \footnotesize \textcopyright 2012 IEEE. Personal use of this material is permitted.
  Permission from IEEE must be obtained for all other uses, in any current or future
  media, including reprinting/republishing this material for advertising or promotional
  purposes, creating new collective works, for resale or redistribution to servers or
  lists, or reuse of any copyrighted component of this work in other works.
  DOI: \href{https://doi.org/10.1109/LRA.2022.3146558}{https://doi.org/10.1109/LRA.2022.3146558}}
\newcommand\copyrightnotice{%
\begin{tikzpicture}[remember picture,overlay]
\node[anchor=south,yshift=10pt] at (current page.south) {\fbox{\parbox{\dimexpr\textwidth-\fboxsep-\fboxrule\relax}{\copyrighttext}}};
\end{tikzpicture}%
}
\begin{document}
\title{Online Next-Best-View Planner for 3D-Exploration and Inspection With a Mobile Manipulator Robot }
\author{Menaka Naazare \orcidA{}, Francisco Garcia Rosas \orcidB{}, and Dirk Schulz \orcidC{}, \textit{Member, IEEE }
\thanks{Manuscript received September 9, 2021; accepted January 7, 2022. Date of
publication January 27, 2022; Date of current version February 15, 2022. This
letter was recommended for publication by Associate Editor M. Popovi{\'c} and
Editor J. Civera upon evaluation of the reviewers’ comments. \textit{(Corresponding
author: Menaka Naazare.)}}
\thanks{The authors are with the Department of Cognitive Mobile Systems (CMS), Fraunhofer FKIE, 53343 Wachtberg, Germany{\tt\footnotesize \{menaka.naazare, francisco.garcia.rosas, dirk.schulz\}@fkie.fraunhofer.de}}
\thanks{This letter has supplementary downloadable material available at \href{https://youtu.be/nsJ_LCio0h0}{https://youtu.be/nsJ\_LCio0h0} and the source code at \href{https://github.com/fkie/fkie-nbv-planner}{https://github.com/fkie/fkie-nbv-planner}, provided by the authors. }

}

\markboth{IEEE Robotics and Automation Letters, VOL. 7, No. 2, April 2022}
{Naazare \MakeLowercase{\textit{et al.}}: Online WG-NBVP for 3D-Exploration and Inspection with a Mobile Manipulator Robot}

\maketitle
\quad
\copyrightnotice
\begin{abstract}
Robotic systems performing end-user oriented autonomous exploration can be deployed in different scenarios which not only require mapping but also simultaneous inspection of regions of interest for the end-user.
In this work, we propose a novel Next-Best-View (NBV) planner which can perform full exploration and user-oriented exploration with inspection of the regions of interest using a mobile manipulator robot.
We address the exploration-inspection problem as an instance of Multi-Objective Optimization (MOO) and propose a weighted-sum-based information gain function for computing NBVs for the RGB-D camera mounted on the arm.
For both types of exploration tasks, we compare our approach with an existing state-of-the-art exploration method as the baseline and demonstrate our improvements in terms of total volume mapped and lower computational requirements.
The real experiments with a mobile manipulator robot demonstrate the practicability and effectiveness of our approach outdoors. 
\end{abstract}


\section{INTRODUCTION}
\IEEEPARstart{D}{uring} applications like disaster response or industrial facility inspection, experts are required to enter potentially hazardous places, e.g., to remove toxic substances, repair damaged machinery or rescue casualties. 
Using robots to investigate the environment beforehand, can greatly improve the situational awareness and help to minimize exposure to contaminants for the experts.
This initial unmanned assessment has to be carried out fast in many applications, which demands an efficient automated exploration approach that should be able to focus on mission relevant aspects such as closely inspecting detected areas of contamination. 

In this work, we propose a Next-Best-View planning approach for this purpose that continuously computes view poses for the robot which promise the most information gain.
Our planning algorithm is specifically tailored for the use of Unmanned Ground Vehicles (UGVs) equipped with a manipulator arm for positioning the sensor. 
Unmanned Aerial Vehicles (UAVs) are a popular platform for this task due their agility and speed, but recent research has already demonstrated the ability of UGVs to monitor and gather data in applications such as disaster response~\cite{yokokohji2021use}, precision agriculture~\cite{gonzalez2020unmanned} and mining~\cite{miura2018field} as well.
UGVs offer the special benefit of being able to carry heavier sensor payloads allowing to take a variety of measurements simultaneously and they can keep the sensors' position steady for longer periods of time.
The additional use of a manipulator arm can enable them to measure in narrow areas, difficult to reach using UAVs.
\begin{figure} [!tbp]
		\includegraphics[width=0.97\linewidth]{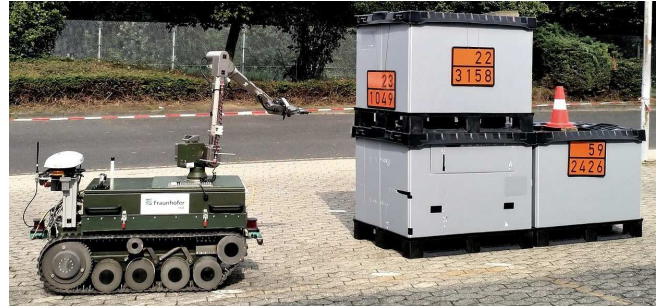} 
		\includegraphics[width=0.97\linewidth]{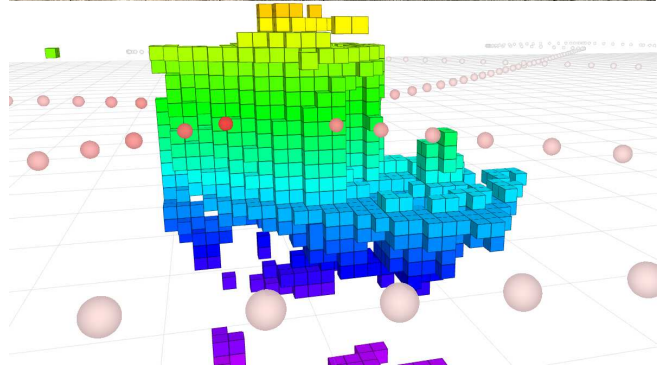}  
	\caption{Our mobile manipulator robot executing the NBV goals generated by our WG-NBVP to perform autonomous exploration and inspection around the plastic pallet boxes (top). The resulting OctoMap~\cite{hornung2013octomap} (bottom) can be visualized online as the robot explores and inspects regions with high contamination levels (depicted as red circles).
	}
	\label{fig:teaser}
\end{figure}

The task of collecting extensive sensor information about an environment efficiently requires to plan paths for the robot that maximize the information gain about the unknown environment.
This problem is also known as informative path planning (IPP). 
Among the solutions to the IPP problem, sampling-based approaches have been successfully applied in closely related applications like
surface reconstruction~\cite{hardouin2020surface}, volumetric mapping~\cite{bircher2016receding} and 3D reconstruction~\cite{vasquez2014view}.
The general idea of the sampling-based IPP solution is to iteratively expand a Rapidly Exploring Random Tree (RRT)~\cite{lavalle1998rapidly} of reachable robot configurations starting from the robot's current pose, and then to move the robot to the first configuration on the best branch. 
Here, the quality of a branch is determined based on the expected information gain at the candidate configurations it contains.
Typically, the branch containing the candidate configuration with highest expected gain is chosen.

Our solution is based on this approach. 
However, directly planning in the joined configuration space of the robot's pose and the manipulator's joint state would lead to plans
which are inefficient to execute.
The information gain of a configuration solely depends on the sensor's pose in 3D space, 
and typically a reachable sensor pose can be attained by a large set of robot configurations, 
most of which require robot platform and manipulator motion.
Therefore, our algorithm plans in the space of sensor poses and determines NBVs instead. 
In addition, it prioritizes manipulator motion over platform motion, as they are generally faster to execute.
The algorithm only decides to drive, if the sampled sensor pose with maximum gain is no longer reachable by manipulator motion alone. 

This work aims at applications in the context of hazardous substance inspection during disaster response, i.e., in chemical or nuclear plants where it is of interest to obtain a situational picture of contaminated areas.
The goal of the approach is to acquire a volumetric map of these areas efficiently by taking measurements of radiation or chemical contamination into account. 
It is assumed that these measurements have been already obtained either with a mobile robot or manually from humans earlier on during a first quick examination.
We call this application exploration-inspection, because it specifically aims at continuously taking closer look at the regions of interest (ROI).
The idea is to use the given contamination information to guide the view point selection for a depth camera during exploration.
Therefore, a contamination intensity map is employed to focus the exploration on regions where the contamination is high. 
This is achieved by taking the intensity map into account in the expected gain estimation used for view point selection.

The main contribution of this work is a novel online NBV planner for a mobile manipulator robot to execute exploration-inspection missions in the context of disaster response.
We name this planner WG-NBVP, because it facilitates a weighted gain function to decide between exploring more space and closer inspection.
To evaluate our approach, we compare it with an existing state-of-the-art method as the baseline.
We conduct simulation and real world experiments using a mobile robot equipped with a 7 degrees of freedom manipulator arm. 
This setup is illustrated in Fig~\ref{fig:teaser}. We demonstrate that our approach reliably explores and shows improved performance not only for our exploration-inspection problem, but also for plain exploration tasks.
Our work builds upon the state-of-the-art Receding Horizon Next-Best-View Planner (RH-NBVP)~\cite{bircher2016receding} and Autonomous Exploration Planner (AEP)~\cite{selin2019efficient} and we show that our approach improves upon aspects such as total explored volume and lower computational requirements.
Finally, WG-NBVP\footnote{[Online]. Available: https://github.com/fkie/fkie-nbv-planner} and the simulation scenarios are released as open source for future research.

\section{Related work}
\noindent Recent solution trends towards exploration problems are either frontier-based methods~\cite{yamauchi1997frontier}, sampling-based methods~\cite{choset2005principles} or a combination of both.
In frontier-based methods, the exploration progresses by driving towards the frontiers, which are regions bordering between the known and the unknown space.
Typically, a utility function is employed to iteratively determine the best frontier until there are no more frontiers left in the environment. 
On the other hand, sampling-based methods use single or multi-query planners to search through the environment~\cite{bircher2016receding,selin2019efficient,hauser2015lazy} and generate view configurations by growing a tree-like data-structure.
Exploration progresses by iteratively moving to the NBV configuration with highest utility.
In the rest of this section, we review the state-of-the-art methods for autonomous exploration with particular focus on NBV planning for 3D exploration using RGB-D cameras.

The general idea in volumetric approaches is to estimate the total volume of unmapped space that can be mapped when the camera is positioned at a given configuration. 
Bircher et al.~\cite{bircher2016receding} introduced Receding Horizon Next-Best-View Planner (RH-NBVP) where the view points were generated by expanding RRTs~\cite{lavalle1998rapidly} and executing the first node of the best branch in a receding horizon fashion.  
To escape the local minima in RH-NBVP, Witting et al.~\cite{witting2018history} store a library of previous RRT nodes to seed the RRT. 
Similarly, the Autonomous Exploration Planner (AEP)~\cite{selin2019efficient} uses RH-NBVP as the local planning strategy and frontier-based planning as the global strategy to escape the local minima.
Popular utility functions used in these approaches are either exponential~\cite{selin2019efficient} or linear formulations~\cite{corah2019communication}. 
Schmid et al.~\cite{schmid2020efficient} argue that these formulations suffer from the fact that they are strictly increasing and therefore, choosing long sub-trees with sub-optimal nodes. They use the notion of efficiency as the gain formulation for accurate Truncated Signed Distance Function (TSDF) reconstruction and expand a single RRT*~\cite{karaman2011sampling} tree.
Likewise, Hardouin et al.~\cite{hardouin2020surface} used  Lazy Probabilistic RoadMap (PRM)*~\cite{hauser2015lazy} to generate paths to the cluster of viewpoint configurations for the surface inspection problem.
However, most of the TSDF-based methods~\cite{schmid2020efficient, song2018surface,hardouin2020surface} require accurate position estimates for the robot, which makes it challenging to be used on robotic systems with high localization uncertainty. 
Therefore, as an alternative we grow RRTs locally and sample intelligently to reduce the computation costs occurring from repeated gain calculations. 

IPP approaches~\cite{popovic2020informative, hitz2017adaptive} that aim to collect data based on ROIs similar to our exploration-inspection problem have also been presented.
These approaches generate a single optimized trajectory for the robot  to drive over the ROIs once efficiently. 
On the contrary, our goal is to enable the robot to carry out a thorough assessment intended 
for extensive autonomous data collection potentially for a longer period of time. 
Therefore, our aim is not to drive over a single trajectory that connects areas with high 
contamination measurements once, but to take NBV measurements repeatedly from varying distance while focusing on the areas with high levels of contamination at all times.

Majority of the aforementioned approaches were validated using UAVs, but they cannot be transferred directly to generate efficient plans for the configuration space of a mobile manipulator robot. 
Therefore, NBV algorithms that focus on generating viewpoints for the depth sensor mounted on the arm of the mobile robot have also been researched.
Vasquez-Gomez et al. ~\cite{vasquez2014view} proposed an NBV algorithm for reconstruction using a mobile manipulator robot with an ``eye-in-hand" sensor. 
To handle unknown objects, Isler et al.~\cite{isler2016information} proposed an information gain-based NBV algorithm where they assume that the spatial bounds of the objects are known and execute repeated circular motions with the arm from the best view points. 
To reconstruct objects without any prior knowledge, Daudelin and Campbell~\cite{daudelin2017adaptable} proposed a probabilistic approach that provides NBV poses for the camera. 
They demonstrated the adaptability of their approach to reconstruct larger objects in simulation using a vehicle model. However, their real experiments were conducted in a smaller area with shorter objects such as a low shelf with a traffic cone and books underneath it. 

To the best of our knowledge, there seems to be a general lack of approaches which aim at NBV exploration of large areas using mobile manipulator robots.
Therefore, we propose an NBV algorithm to autonomously execute 3D exploration missions using a mobile manipulator robot in partially known and unknown environments. 

\section{Proposed Approach}
\begin{algorithm}[!t]
\caption{{Explore-inspect }}
\label{alg1}
\small
\begin{algorithmic}[1]
\STATE $\text{\bf Inputs:}\: \mathbf{q}_0 \in\mathbf{SE}(3),\, \mathcal{I} = \mathrm{voxel\_ map}\{m_1,\ldots,m_{N_m}\}$
\STATE $\text{\bf Output:}\: \mathcal{M}_T\, \text{\em\,\, // the final volumetric map}$
\STATE $\mathcal{M}_0 = \emptyset$
\STATE $\mathbf{Q}_0 = \{\mathbf{q}_0\}$
\STATE $t \leftarrow 0$
\STATE $\mathrm{finished} \leftarrow \mathbf{false}$
\WHILE{$\neg\mathrm{finished}$}
\STATE $\mathbf{D}_t = \{D = \text{\em capture}(\mathbf{q})\mid \mathbf{q}\in\mathbf{Q}_t\}$
\STATE $\mathcal{M}_{t+1} = \text{\em octomap}(\mathbf{Q}_t, \mathbf{D}_t, \mathcal{M}_t )$
\STATE $\mathbf{\hat{Q}}_t = \cup\, \mathbf{Q}_i\,\,\, 0\leq i\leq t$
\STATE $(\mathcal{V}_t,\mathcal{E}_t) = \text{\em rrt}(\mathbf{q}_t,\mathcal{M}_{t+1},\mathcal{I})$
\STATE $\mathbf{q}_{t+1} = \argmax\limits_{\mathbf{q}\in\mathcal{V}_t} G(\mathbf{q},\mathbf{\hat{Q}}_t,\mathcal{M}_{t+1},\mathcal{I})$
\IF{$g_{\text{\em min}} \leq G(\mathbf{q}_{t+1},\mathbf{\hat{Q}}_t,\mathcal{M}_{t+1},\mathcal{I})$}
\STATE $\mathbf{R}_{t+1} = \text{\em reachable-prefix}(\mathbf{q}_t,\mathbf{q}_{t+1},\mathbf{\hat{Q}}_t,\mathcal{V}_t,\mathcal{E}_t,\mathcal{M}_{t+1}) $
\IF{$\mathbf{R}_{t+1}\neq\emptyset$}
\STATE $\mathbf{Q}_{t+1} = \mathbf{R}_{t+1}$
\ELSE
\STATE $\mathbf{Q}_{t+1} = \{\text{\em approach}(\mathbf{q}_{t+1},\mathcal{M}_{t+1})\}$
\ENDIF
\ELSE
\STATE $\mathrm{finished} \leftarrow \mathbf{true}$
\ENDIF
\STATE $t \leftarrow t + 1$
\ENDWHILE
\STATE $\textbf{return}\, \mathcal{M}_t$
\end{algorithmic}
\end{algorithm}
%
%
\noindent In the context of this work, the goal of 3D exploration is to obtain a 
volumetric map of a ROI within an unknown environment using a mobile 
manipulator robot equipped with a depth camera.
Here, the relevance of an area is determined based on readings $m_i$, $1\leq i\leq N_m$ of a sensor that provides single scalar values like, e.g., a radiation sensor.
These measurements are integrated beforehand into a voxel map $\mathcal{I}$, 
which is used by the algorithm to obtain utility values for the camera 
positions and orientations.
Volumetric mapping is based on depth images $\mathbf{D}_t$ of the camera which 
is assumed to be mounted close to the  end-effector (EEF) of the manipulator.
The task of the WG-NBVP is to provide NBV poses for the depth camera 
that facilitate building the volumetric map efficiently. 

An outline of the exploration-inspection algorithm is given in Algorithm~\ref{alg1}, the NBV computation is summarized in lines 11 to 19. 
In each iteration, the algorithm visits a sequence of camera poses 
$\mathbf{Q}_t$ computed in the previous iteration, captures depth images at 
these locations and integrates them into the volumetric map (lines 8 and 9 of 
Algorithm~\ref{alg1}).
The algorithm then grows an RRT~\cite{lavalle1998rapidly} of possible future 
camera poses into the free space of the current map $\mathcal{M}_{t+1}$, starting from the 
current actual camera position $\mathbf{q}_t$ (line 11). The branches of this 
RRT correspond to sequences of camera poses in the surrounding of the robot.
If the expected information gain of the best pose $\mathbf{q}_{t+1}$ contained in the tree exceeds a lower limit $g_{\text{\em min}}$, the algorithm extracts the poses on the branch to $\mathbf{q}_{t+1}$. The poses on this branch which can be reached with the manipulator arm form the sequence of NBVs $\mathbf{Q}_{t+1}$ to be used in the next iteration (lines 13 and 16).
If none of the poses on the branch to the best node can be reached with the 
manipulator without moving the robot, i.e., if  $\mathbf{R}_{t+1}$ is empty, the 
robot navigates towards the best pose $\mathbf{q}_{t+1}$ (line 18) and the next 
iteration of the algorithm continues from there.
The algorithm finishes if the RRT computation no longer yields any new camera 
poses above $g_{\text{\em min}}$ (lines 13 and 21).
\subsection{RRT Expansion}
\noindent To determine the NBVs, the algorithm maintains an RRT of $N_{\text{\em max}}$ 
camera poses $\mathcal{V}_t$ contained in the free space of the map 
$\mathcal{M}_{t+1}$ explored so far. 
For escaping local minima, similar to approaches \cite{witting2018history} \cite{selin2019efficient}, we store the corresponding nodes of the expanded RRT 
as cached nodes for future use.
In each iteration, the expected gain of all cached nodes exceeding 
$g_{\text{\em min}}$ is reevaluated. 
Only the top candidates are kept and the corresponding lowest gain value is 
assigned as $g_{\text{\em min}}$ for the next iteration. 


The RRT expansion procedure generates random pose samples 
$\mathbf{q}_{\text{\em rand}}$ on an imaginary sphere centered at the base of 
the manipulator arm and with a sampling radius exceeding the maximum reach of 
the arm.
In each iteration, $N$ tries are allowed to sample a $\mathbf{q}_{\text{\em 
rand}}$ and find a $\mathbf{q}_{\text{\em new}}$.
To reduce the dimensionality of the search space, only the position of a camera pose is randomly sampled and the orientation is determined either by computing the positive gradient direction in $\mathcal{I}$ where the measured contamination values are guaranteed to be high or by computing the direction that promises maximum unmapped volume at the sampled position. 
In this way, we ensure that the camera always faces the direction of high utility for a given position based on the mode of operation.
Using the position of $\mathbf{q}_{\text{\em rand}}$, the tree expansion 
determines the closest node $\mathbf{q}_{\text{\em near}}$ of the RRT and 
computes the position of a new node $\mathbf{q}_{\text{\em new}}$ on the 
straight line connection between $\mathbf{q}_{\text{\em rand}}$ and 
$\mathbf{q}_{\text{\em near}}$ at the step size $l$ from 
$\mathbf{q}_{\text{\em rand}}$. 
$\mathbf{q}_{\text{\em new}}$ is added to the tree only if its straight 
line connection to $\mathbf{q}_{\text{\em near}}$ is in free space.
Additionally, to avoid redundant view configurations, 
$\mathbf{q}_{\text{\em new}}$ is added to the RRT only if it is within the 
exploration bounds and at minimum sampling 
distance $d$ $<$ $l$ to the existing nodes of the RRT.

\subsection{Weighted-Sum Information Gain}
\noindent The information gain function $G$ employed by the algorithm combines contributions of three objectives: 1) to explore novel space ($G_f$), 2) to prioritise relevant areas ($G_m$), and 3) to avoid redundant observations ($G_v$). 
For this purpose a weighted-sum objective~\cite{caramia2008multi,jakob2014pareto,marler2010weighted} is used that combines the individual objectives into a single scalar function. 
The function determines the utility of a pose $\mathbf{q}$ given the current map $\mathcal{M}_{t}$, and the voxel map $\mathcal{I}$,
\begin{equation}\label{eqn:ig}
G(\mathbf{q}, \mathcal{M}, \mathcal{I}) = w_{f} G_{f} +  w_{m} G_{m} + w_{v} G_{v}.
\end{equation}
Here, $w_{f}$, $w_{m}$,  $w_{v}$ are the weights assigned to control the contribution of the individual objectives. They are defined for a given camera pose $\mathbf{q}$ as follows:
\begin{enumerate}
\item Gain free space $G_{f}$: The volume of the unmapped space that can be 
covered within the horizontal and vertical fields of view (FoV) of the RGB-D 
camera if it were at that position. We use the volumetric gain equations as 
given in \cite[Section~VI]{selin2019efficient}. For the generated map $M$, the 
gain of the volume element $dV$ at radius $r$ from the sensor in the direction 
$(\theta, \phi)$ with dimensions $(\Delta_{r}, \Delta_{\theta}, \Delta_{\phi})$ 
is given by:
\begin{equation} \label{g_dv}	
g_{dV}(r, \theta, \phi) =
	\left\{ \begin{array}{ll}
	dV(r, \theta, \phi) &  \mathcal{M}(r, \theta, \phi) \; \text{is unknown} \\ 
	0 & \text{otherwise}.
	\end{array} \right.
	\end{equation}
where
\begin{equation}
	dV(r, \theta, \phi) = \left (2r^2\Delta_{r} + \frac{1}{6}\Delta_{r}^{^{3}}   \right )\Delta _{\theta}sin(\phi)sin(\Delta _{\phi}/2).
\end{equation}
Then, for the given $\psi$, the total free space gain is:
		\begin{equation} \label{g_f}
	G_{f} = \sum _{\theta  = \psi - fov_{\theta}/2}^{\psi + fov_{\theta}/2} \sum _{\phi  = - fov_{\phi}/2}^{fov_{\phi}/2} \sum _{r  = 0}^{\text{max or hit}} 	g_{dV}(r, \theta, \phi).
	\end{equation}
\item Gain measurement $G_{m}$: The scalar sensor measurement value for the voxel corresponding to $\mathbf{q}$, i.e., $G_m = \mathcal{I}(\mathbf{q})$.
\item Gain visited $G_{v}$: The penalty for the planned camera position depending upon whether it was previously assigned,
	\begin{equation} \label{g_v}
G_v=
\left\{ \begin{array}{ll}
-1, &\text{if } \mathbf{q}\in\mathbf{Q}\\
0, &\text{otherwise.}
\end{array} \right.
	\end{equation}
\end{enumerate}	
\subsection{Arm and Robot Motion}
\noindent Based on the current RRT, the Explore-inspect algorithm decides on a sequence 
of NBV poses $\mathbf{Q}_{t+1}$ to visit in the next iteration of the algorithm.
The algorithm prefers to only use the manipulator arm as long as possible, because moving the arm is more efficient and also more precise than moving the robot.
In each iteration, the next best pose $\mathbf{q}_{t+1}$ contained in the RRT is determined.
Even if this pose itself is not reachable with the arm, it might still be possible to gather new information by approaching $\mathbf{q}_{t+1}$  with the arm as far as possible.
For this purpose, the algorithm checks if reachable poses exist on the branch to $\mathbf{q}_{t+1}$ which have not been visited before.
These poses are then chosen as the NBVs $\mathbf{Q}_{t+1}$ to visit.
Here, the reachability tests are based on the inverse kinematics of the 
manipulator arm and collision tests in the current volumetric map 
$\mathcal{M}_{t+1}$.
In case no novel reachable views exist on the branch to $\mathbf{q}_{t+1}$, the algorithm decides to drive the robot. 
For this purpose, the best camera pose $\mathbf{q}_{t+1}$ is projected to the 
ground plane and assigned to the navigation system of the robot.
At the destination, the arm is oriented to face $\mathbf{q}_{t+1}$.
The current implementation relies on the 
teb\_local\_planner\footnote{[Online]. Available:https://wiki.ros.org/teb\_local\_planner} and the 
MoveIt\footnote{[Online]. Available: https://moveit.ros.org/} framework of the Robot Operating 
System (ROS) for robot navigation and to perform motion planning for the arm. 
\subsection{Exploration Mode}
\noindent For a fair comparison of WG-NBVP with NBV approaches for volumetric mapping 
that do not adopt prior information from intensity sensor, we 
additionally implemented a pure exploration version of the algorithm. 
In this exploration mode, the voxel map $\mathcal{I}$ does not exist. 
Accordingly, the second term $G_m$ of the gain function is omitted, when evaluating poses during RRT expansion. 
Furthermore, the computation of the camera orientation during pose sampling needs to be changed, because the gradient of $\mathcal{I}$ is not available.
Instead, the camera is oriented towards the direction that maximises $G_{f}$.
This basically causes the camera to always point towards the horizon and to maximize the amount of unmapped space.

\section{Evaluation}
\noindent We validated WG-NBVP both in simulation experiments and on the real mobile manipulator robot. 
Since AEP~\cite{selin2019efficient} was shown to perform well for both indoor and outdoor environments, we compare our approach to it in full exploration mode and exploration-inspection mode. We use the software implementation provided by the authors to conduct our experiments.

\begin{figure}[!tbp]
\includegraphics[width=0.237\textwidth]{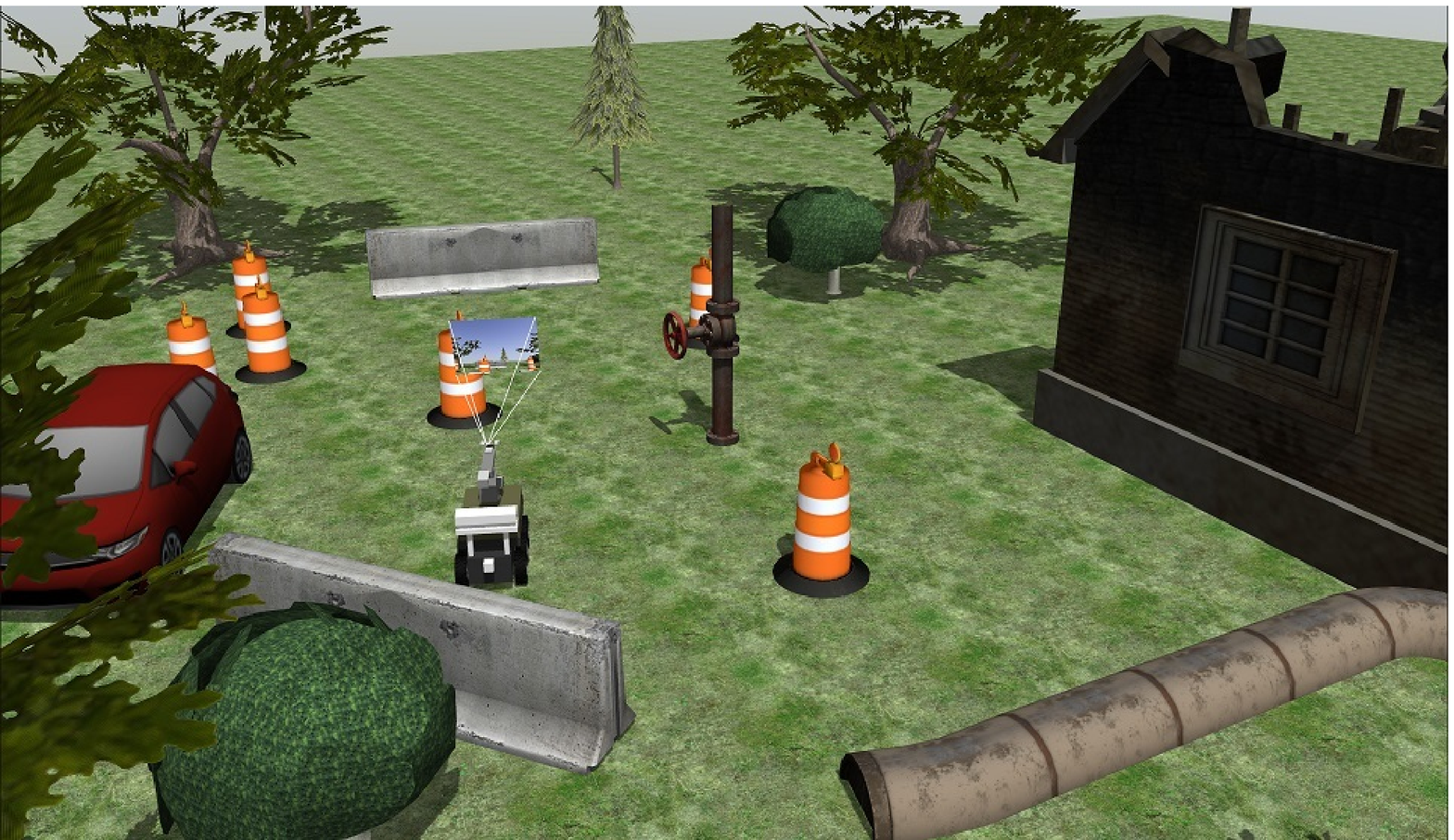}\includegraphics[width=0.24\textwidth]{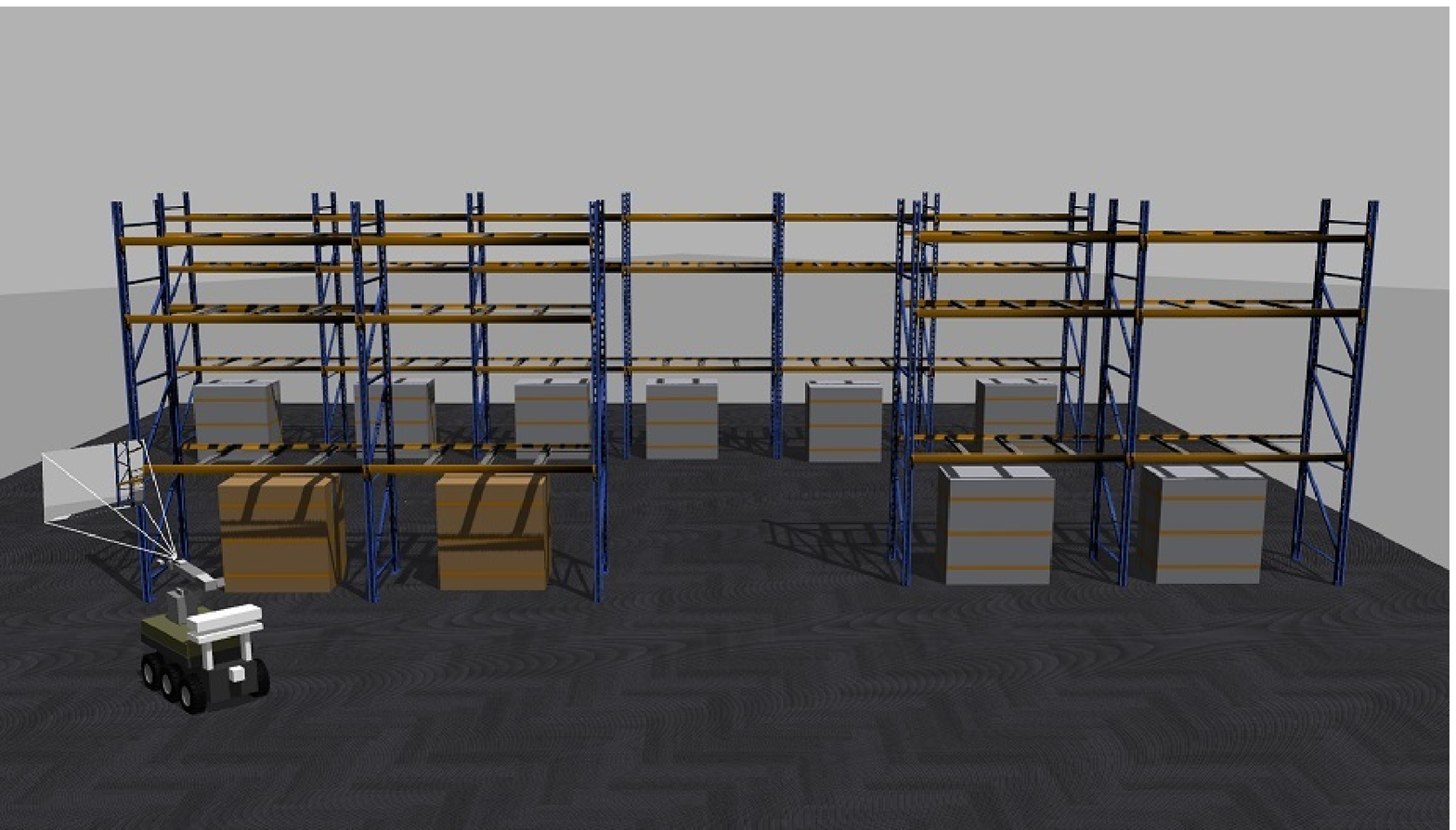}
\caption{Two Gazebo simulation scenarios used for evaluation. For the outdoor disaster scenario (left), the robot has to explore and inspect the area around the valve. Similarly, for the warehouse scenario (right), the robot needs to explore and inspect the area around the shelf with the brown boxes.}
\label{fig:sim_scenarios}
\vspace{-2ex}
\end{figure}

Three different metrics are used to evaluate the performance of the planners: 1) the percentage of ROI explored, which indicates the amount of voxels observed within the ROI where the measured sensor readings $m_i$ have higher values, 2) the percentage of environment explored, which expresses the total mapped volume enclosing the ROI and, finally, 3) the computation time taken to generate plans as the exploration progresses.
Due to the non-deterministic nature of the runs, we repeated the experiments for 10 runs on a workstation with Intel i7-9700 CPU at 3 GHz with 8 cores and 32\,GB of RAM.

\subsection{Experimental Setup}
\begin{figure}[!tbp]
	\centering
	\includegraphics[width=0.4\textwidth]{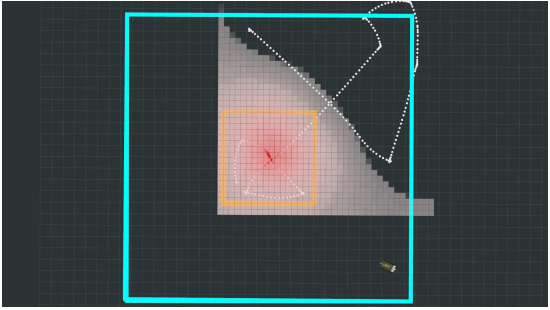}
	\caption{Top down view of the exploration boundary (cyan), the ROI (orange) and prior recorded contamination values $m_i$ interpolated as red and white voxels representing high to low contamination levels.}	
	\label{fig:25m_box_8m_ROI}
	\vspace{-2ex}
\end{figure}

\noindent Realistic simulation scenarios were built using the Gazebo\footnote{[Online]. Available: https://gazebosim.org/} simulator to further demonstrate the performance of the proposed planner in practical settings. 
Therefore, we modelled an outdoor disaster scenario with a valve at the centre and an indoor warehouse scenario with shelves as shown in Fig.~\ref{fig:sim_scenarios}. 
The robot's navigation system relies on 2D laser range sensors for collision avoidance which are mounted at its front and back. 
Both of the environments are quite challenging to navigate based on the provided laser scans, especially due to the gaps caused by the structure of the shelves, short-sized objects such as traffic cones and the narrow spaces between them.
All experiments take place inside an outer exploration boundary of 25\,m  $\times$ 25\,m in size that contains a single ROI of 8\,m  $\times$ 8\,m. 
Fig.~\ref{fig:25m_box_8m_ROI} shows an example of the exploration boundary, the ROI and its underlying voxel map $\mathcal{I}$ built from the measured contamination values $m_i$.

Like the real robot system, the simulated robot uses MoveIt to perform motion planning for the arm and the teb\_local\_planner for the local navigation of the base. 
Our mobile manipulator robot is equipped with an Intel RealSense D435 camera that is mounted near the EEF and provides depth images for building an OctoMap~\cite{hornung2013octomap}. 
The parameters used for perception for the experiments are summarized in Table~\ref{tab:cam_params}.
Except for the GPS system used on the real robot and differences resulting from sensor and controller simulations, the setup of real and simulated robot are very similar.
Only the vertical workspace dimension of the manipulator was limited from  0.4\,m to 1.4\,m in simulation, while a lower limit of 0.6\,m was chosen during the real experiments to prevent self-collisions with additionally mounted cameras.
The maximum reach of the manipulator arm was limited to 1.3\,m for all experiments. 

\begin{table}[!tp]
	\centering
	\caption{\label{tab:cam_params} Perception parameters used for all experiments}
	\begin{tabular}[t]{ll}
		\toprule
		FOV $h_{fov}\times v_{fov}$ & $86^{\circ}\times 57^{\circ}$ \\
		Octomap resolution $r $ & 0.1\,m \\
		Min. range $r_{min} $ & 0.3\,m \\
		Max. range  $r_{max} $ & 1.5\,m \\
		\bottomrule
	\end{tabular}
\end{table}%

Prior to the comparisons, we optimized AEP's parameters, starting with the reference values given in~\cite[Section~VII]{selin2019efficient}.
We tuned the parameters according to the  warehouse scenario since it is the most challenging to navigate.
We finally fixed the parameter values to $g_{zero} = 2.0 $, $\lambda = 0.25$ and $N_{max} = 600$ as these yielded the highest percentage of volume explored compared to other combinations. 
All parameter values used for AEP and WG-NBVP are summarized in Table~\ref{tab:aep_wg_nbvp_params}. 
For WG-NBVP in exploration mode, we interchange the weights for free space and measurements.
Throughout all experiments, the parameters are kept the same for all the planners in all the scenarios unless otherwise specified.
%
\begin{table} [tbp]
\centering
\caption{\label{tab:aep_wg_nbvp_params}Planner parameters for both the scenarios}
\begin{tabular}[t]{lll}
 \toprule
 & AEP & WG-NBVP\\
\midrule
RRT step size $l$ & 0.5\,m & 0.5\,m\\
RRT node max. tries $N$ & 50 & 50\\
RRT max. nodes $N_{max} $ &  600 & 600\\
Min. gain $g_{zero} $ & 2.0 & n.a\\
Degrees coefficient $\lambda $ & 0.25 &  n.a\\
RRT collision radius $r_{b} $ & 0.25\,m & 0.25\,m\\
Weight free space $w_{f} $ &  n.a & 5.0 \\
Weight sensor measurement $w_{m} $ &  n.a & 1.0 \\
Weight visited cell $w_{v} $ &  n.a & 500 \\
\bottomrule
\end{tabular}
\end{table}
\begin{figure*} 

		\includegraphics[width=0.33\textwidth]{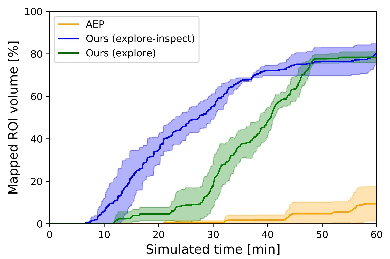}
		\includegraphics[width=0.33\textwidth]{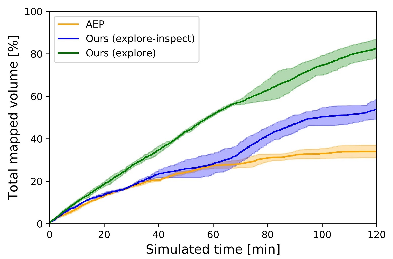} 	
		\includegraphics[width=0.33\textwidth]{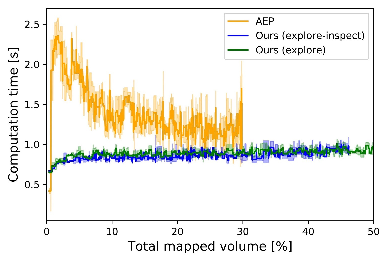} \\

		\includegraphics[width=0.33\textwidth]{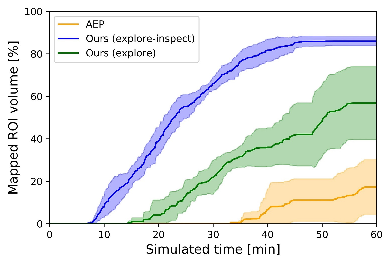}
		\includegraphics[width=0.33\textwidth]{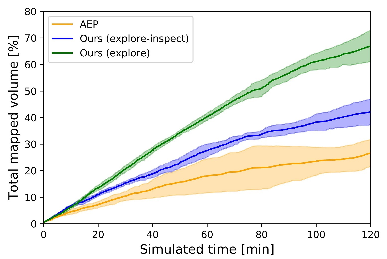}
		\includegraphics[width=0.33\textwidth]{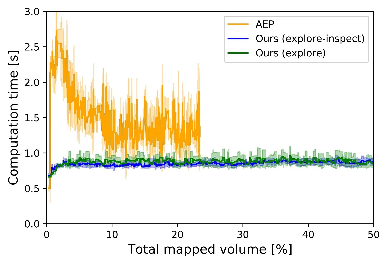} \\
		
	\caption{Results obtained during simulation experiments for the outdoor disaster (top row) and the warehouse scenario (bottom row). The blue curves indicate the results achieved by our approach in explore-inspect mode, the green curves represent our approach in full exploration mode and lastly, the results obtained for AEP~\cite{selin2019efficient} are marked by the orange curves. Mean and standard deviation per approach are shown.}
	\label{fig:sim_plots}
	\vspace{-2ex}
\end{figure*}
\subsection{Results and Discussion}
\begin{figure*}
\begin{tabular}{lcr}
\includegraphics[width=0.3\textwidth,trim={0 2.4cm 0 0},clip]{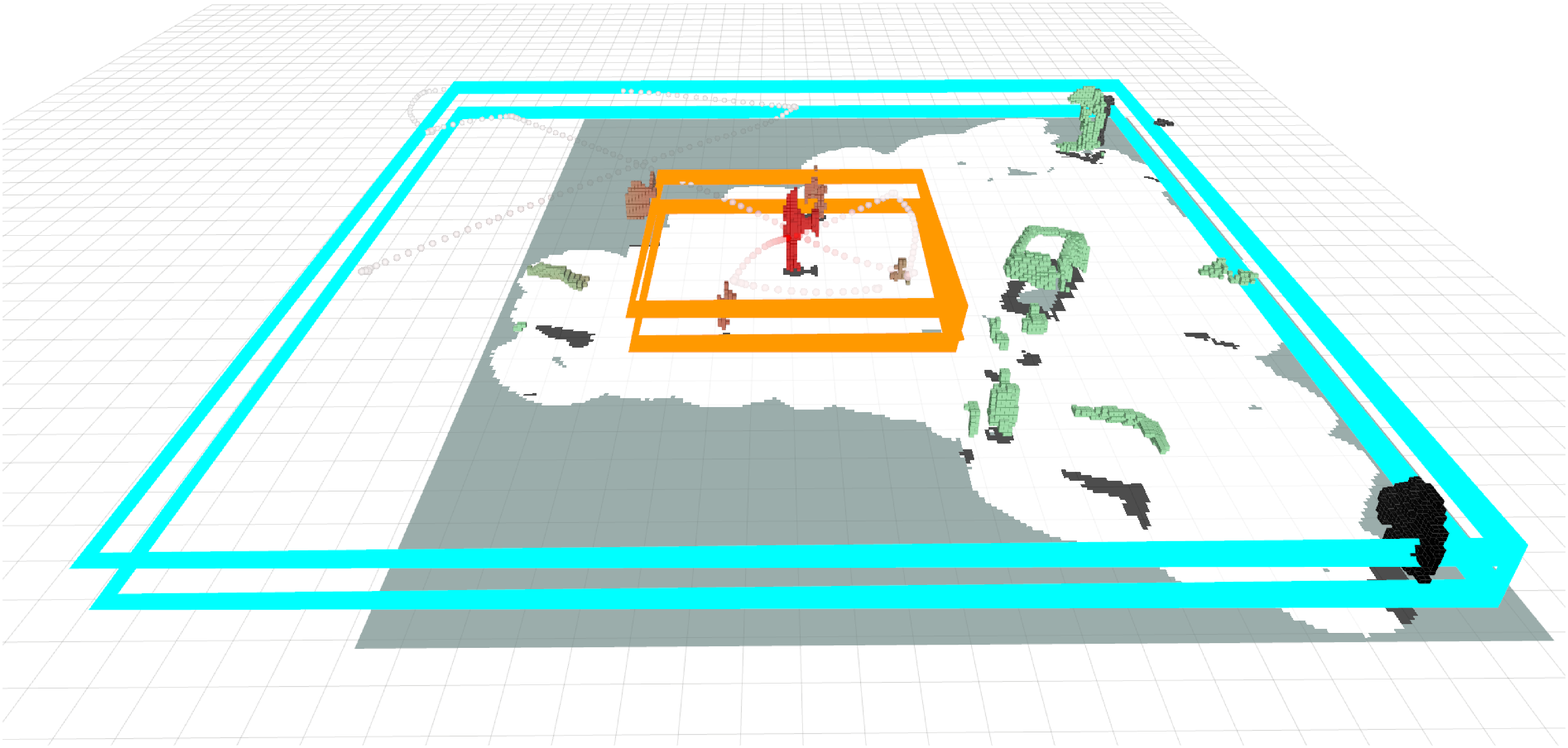}&\includegraphics[width=0.3\textwidth]{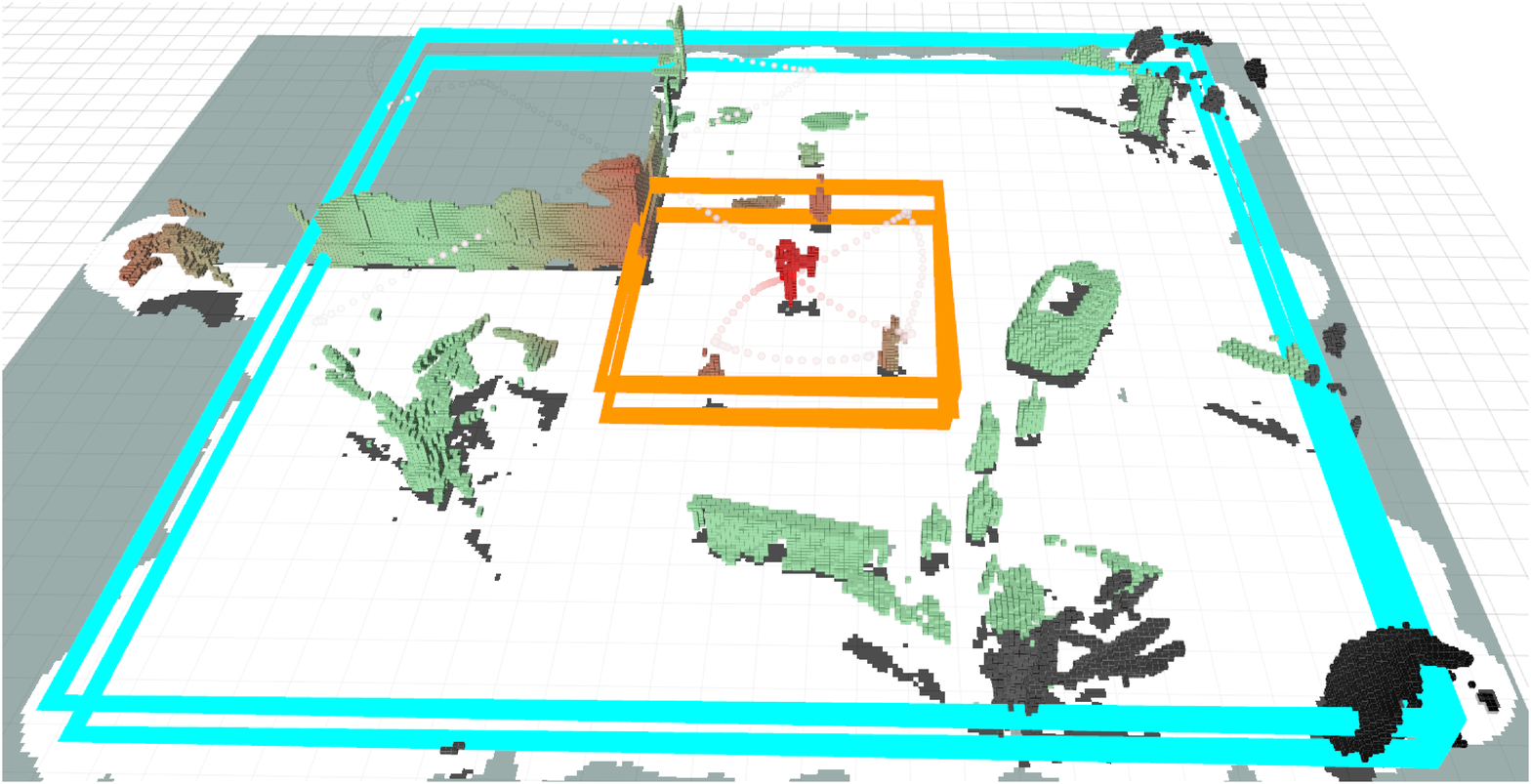}&\includegraphics[width=0.288\textwidth]{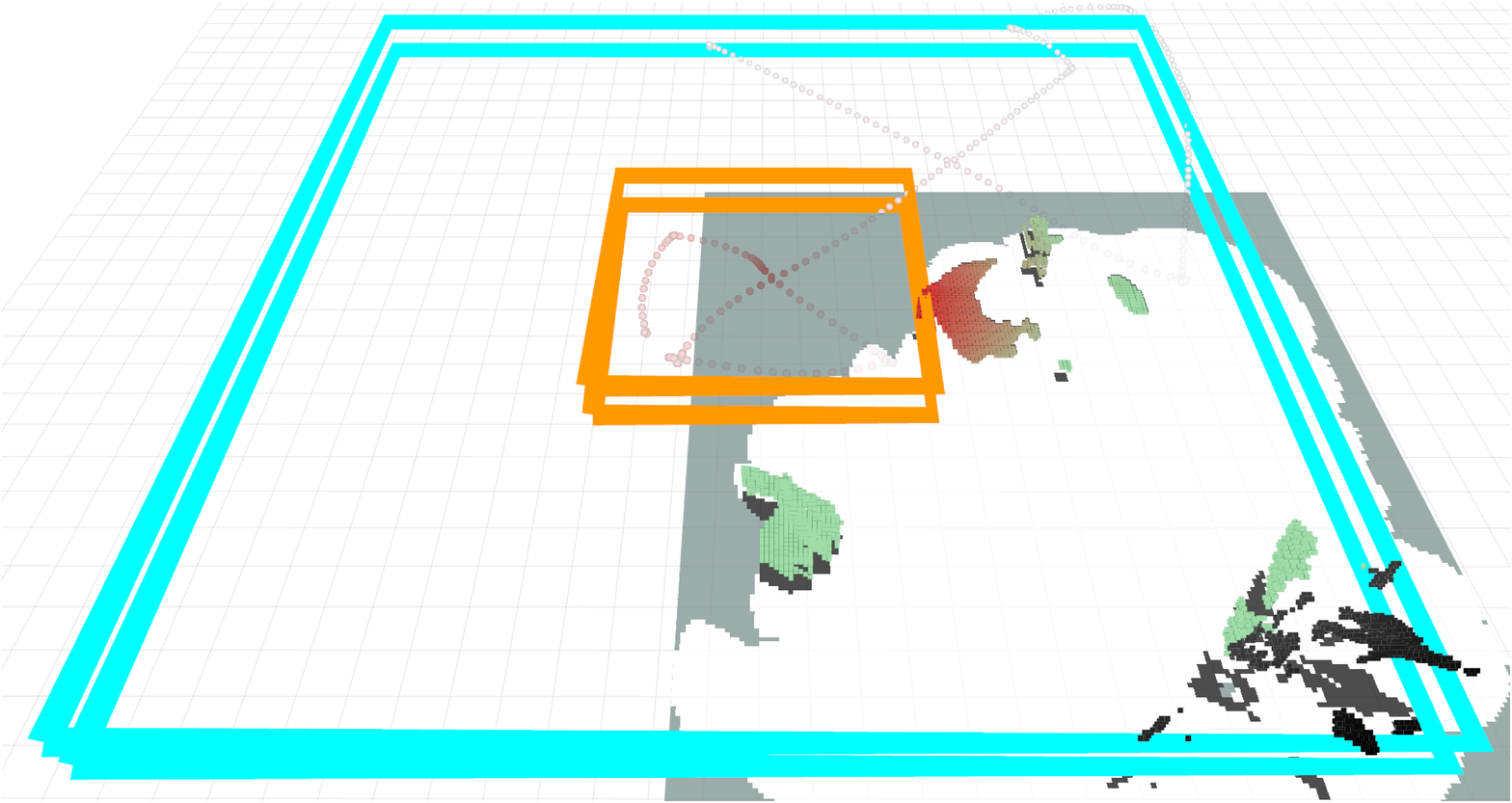}\\ 
\includegraphics[width=0.3\textwidth,trim={0 0 0 0},clip]{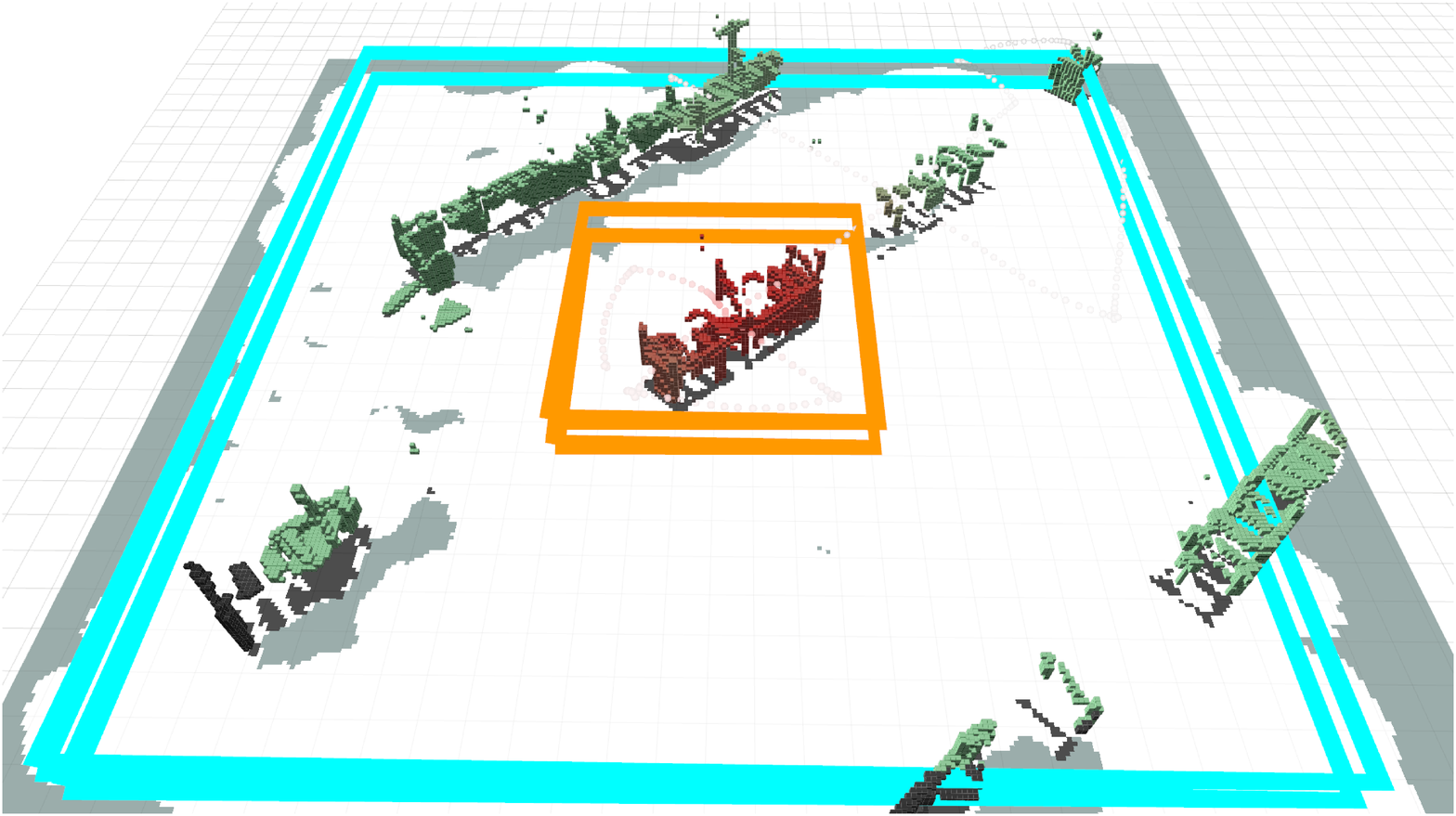} &\includegraphics[width=0.3\textwidth]{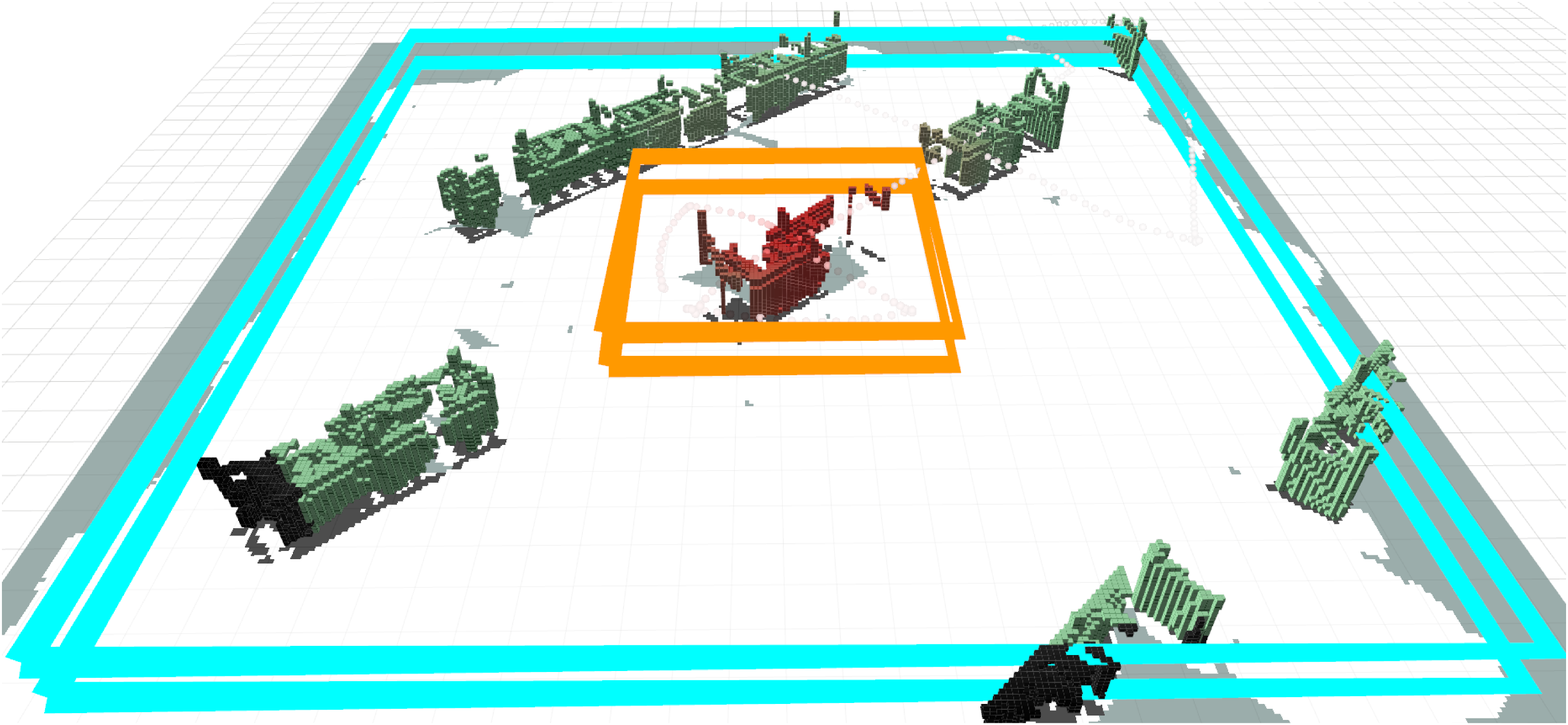} &\includegraphics[width=0.289\textwidth]{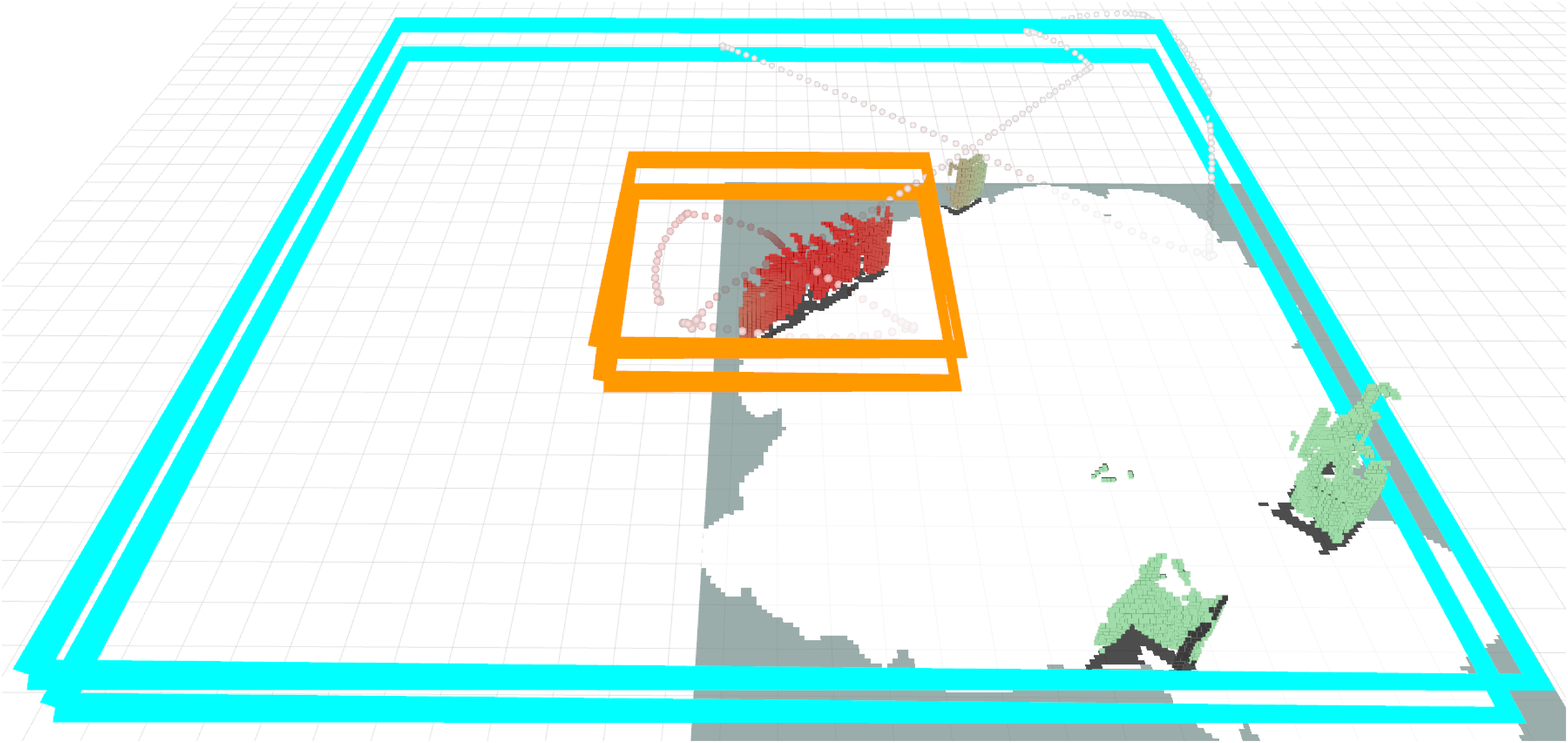}\\ 
\end{tabular}		
\caption{The octomaps generated by the different planners (from the left Ours (explore-inspect), Ours (explore) and AEP) for outdoor disaster (top row) and warehouse scenario (bottom row). Cyan and orange bounding boxes represent the exploration boundary and the ROI respectively. The voxels inside ROI are colored red to represent higher contamination values whereas green voxels indicate lower contamination values and the dark grey voxels imply that the contamination intensity is unknown.}
\label{fig:sim_maps}
\vspace{-2ex}
\end{figure*}

\noindent In order to evaluate the differences in performance between the explore-inspect mode and the exploration mode of WG-NBVP, we first compared the planners on exploring the ROI inside the test environments as fast as possible, followed by a comparison of their performance in case of a full exploration of the environment. 
Fig.~\ref{fig:sim_plots} shows the simulation results obtained for the different planners on different simulated scenarios. 
Within 40 minutes our approach is able to map about 80\% of the ROI (top-left and bottom-left) in all the runs for both scenarios, demonstrating the ability of WG-NBVP to reliably and quickly map areas that are interesting for the experts.
The weighted-sum gain function enables the robot to quickly scan relevant areas where the measured contamination values $m_i$ are high, and to achieve an efficient targeted exploration, this way.  
Being applied on a ground robot, AEP explores less than 40\% of the total environment and maps even lesser amount of the ROI in the same amount of time. 
In this application, AEP suffers from several drawbacks such as failure to escape local minima, longer planning times and frequent back and forth movements when the robot gets stuck in dead-ends. 
In contrast, WG-NBVP using exploration mode maps more than 70\% in both scenarios and requires less than a second of planning time per iteration, exhibiting a superior efficiency of planning for both the arm and the base of the mobile manipulator robot.
Fig.~\ref{fig:sim_plots} (top-right and bottom-right) shows the computation time versus the total volume mapped.
It can be observed that each planning request is answered at least two times faster by WG-NBVP than by AEP on average. 
This is achieved by thoroughly refining the NBV poses in each iteration to avoid generating a cluster of similar view poses and by limiting the cache-node capacity to a small size of high utility nodes. 
On the contrary, AEP continuously updates the gain of cluster of similar nodes that lie above a certain threshold and within a given radius of the robot. 
The maps generated from the simulation experiments for the outdoor disaster scenario (top) and the warehouse scenario (bottom) are shown in Fig.~\ref{fig:sim_maps}. 
\subsection{Impact of Proposed Changes}
\noindent In the second set of experiments, we study the impact of the gain threshold computation, RRT sample generation, and the cache node filtering technique on the performance of the explore-inspect algorithm.
To evaluate the contribution of the variable gain threshold computation, we carried out sets experiments with variable and fixed gain threshold. 
These are denoted as VG and FG in the results shown in Table~\ref{tab:ablation_components}.
We compare the sampling strategy of WG-NBVP to the strategy used by AEP~\cite{selin2019efficient}, which are labelled WGS and AEPS respectively. 
Lastly, we compare the contribution of the cached nodes filtering technique of WG-NBVP to the filtering method used by AEP, labelled as CF and NCF. 

Table~\ref{tab:ablation_components} shows the results obtained using these 8 different settings in the explore-inspect algorithm for the outdoor disaster scenario.
The percentage of space explored within 60 minutes and the computation time required to map 50\% of the ROI's volume are reported. 
It can be clearly noticed that combinations with cached nodes filtering (CF) and storing only a limited number of high utility nodes leads to reduced computation time in almost every case. 
While gain updates for the cached nodes around the robot (NCF) lead to higher computation due to a larger number of nodes and possibly redundant views. 
With cache node filtering, our sampling strategy (WGS) combination achieves much better ROI exploration irrespective of the type of 
gain threshold due to uniformly sampling around the hemisphere and adding a $\mathbf{q}_{\text{\em new}}$ only if it is not too close to 
any other samples of the current RRT. 
On the contrary, when no cache node filtering (NCF) is enabled, AEPS with variable gain shows better performance as the generated RRTs spread larger parts of the environment whereas our sampling technique tends to generate dense smaller RRTs limited to the size of the sphere causing slower exploration. 
Adding a variable gain (VG) further increases the amount of ROI explored.
In each iteration, the choice of the best node $\mathbf{q}_{t+1}$ depends on the lowest expected gain $g_{\text{\em min}}$ found in the set of cached nodes. 
This introduces a stricter threshold for the decision on when to drive to new positions and allows the robot to escape local minima. 
In contrast, fixed gain AEP sampling and filtering significantly limit the amount of the ROI explored and cause higher computation costs.
   
\begin{table} [!tbp]
\centering
\caption{\label{tab:ablation_components}  Algorithm\;\ref{alg1} ablation study }
\begin{tabular}[t]{lcccc}
 \toprule
  & ROI &  Exploration & Computation time\\
  &60 min [\%]  & 60 min [\%] &  50\% ROI [s]\\
\midrule
FG, AEPS, NCF & 48.3 $\pm$ 3.1 & 30.5 $\pm$ 3.7 & 2.0 $\pm$ 0.0\\
FG, WGS, NCF & 55.1 $\pm$ 9.1 & 28.9 $\pm$ 1.7& 5.2 $\pm$ 1.3\\
FG, AEPS, CF & 55.6 $\pm$ 3.9 & 25.5 $\pm$ 6.0 & \textbf{0.5 $\pm$ 0.1}\\
FG, WGS, CF & 73.9 $\pm$ 3.5 & \textbf{31.3 $\pm$ 4.6} & 0.8 $\pm$ 0.0\\
VG, AEPS, NCF & 64.9 $\pm$ 4.3 & 20.9 $\pm$ 0.1 & 3.2 $\pm$ 0.9\\
VG, WGS, NCF & 55.3 $\pm$ 2.6 & 17.8 $\pm$ 3.0& 7.7 $\pm$ 1.0\\
VG, AEPS, CF & 70.7 $\pm$ 3.6 & 30.4 $\pm$ 3.6& \textbf{0.5 $\pm$ 0.0}\\
\textbf{VG, WGS, CF}  & \textbf{82.6 $\pm$ 5.3} & 28.8 $\pm$ 4.0 & 0.9 $\pm$ 0.0\\

\bottomrule
\end{tabular}
\end{table}
\begin{table} [!bp]
\centering
\caption{\label{tab:gain_functions} Comparison of utility functions}
\begin{tabular}[t]{lcccc}
 \toprule
  & ROI &  Exploration & Distance travelled\\
  &60 min [\%]  & 60 min [\%] & 60 min [m]\\
\midrule
Exponential & 82.0 $\pm$ 2.7 &\textbf{ 34.0 $\pm$ 8.1} & 184.6 $\pm$ 69.2\\
Linear & 74.4 $\pm$ 2.5 & 28.6 $\pm$ 0.5& 186.8 $\pm$ 42.6\\
\textbf{Weighted-gain} & \textbf{82.6 $\pm$ 5.3} & 28.8 $\pm$ 4.0 & \textbf{167.1 $\pm$ 26.7} \\
\bottomrule
\end{tabular}
\end{table}
\subsection{Influence of Different Utility Functions}
\noindent In contrast to related approaches like the AEP, the WG-NBVP does not take motion costs into account in the evaluation of the NBV.
Instead, the weighted-gain function (\ref{eqn:ig}) is applied without discounting any path costs. Reasonable motion costs are hard to
determine in the application on the mobile manipulator robot, because it uses two qualitatively different modes, arm motion and driving, and node distances in the RRT are only loosely
related to actual execution times of arm motions.
For this reason, the consideration of path costs based on the RRT can indeed lead to inferior results.
To illustrate this, we carried out an experiment where we compare NBV selection based on the weighted gain function against utility functions that penalize motion costs by using 1) an exponential discounting factor~\cite{selin2019efficient}, and 2) a linear penalty term~\cite{corah2019communication}, both depending on the Euclidean distance of a node to its parent in the RRT. 
In both cases we used a proportionality factor of 0.25 times this distance.
The results are presented in Table~\ref{tab:gain_functions}. While the utility function with exponential discounting explores a similar amount of the ROI's  volume within 60 minutes, it tends to drive more.  The utility function using linear path cost penalties shows inferior performance on both metrics. 

\subsection{Robotic Platform Studies}
\noindent The aim of the next experiment was to illustrate the advantage of favoring arm motion over driving the robot.
For this purpose we carried out the experiments with a stationary arm and with planned arm motion.    
In the former case, the arm is kept in a fixed position throughout the mission with the depth camera facing to the front.
The robot is driven to each chosen NBV goal.
Table~\ref{tab:mobile_arm} shows the amount of ROI explored versus the distance travelled over 60 minutes for both cases.
It can be observed that planned arm motion offers significant improvements in terms of ROI explored and the total distance travelled. 
The mobile arm allows the depth camera to gather information from a large number of view points without the need to move the platform.
\begin{table} [!tp]
\centering
\caption{\label{tab:mobile_arm} Impact of mobile manipulator motion}
\begin{tabular}[t]{lcccc}
 \toprule
  & ROI &  Exploration & Distance travelled\\
  &60 min [\%]  & 60 min [\%] & 60 min [m]\\
\midrule
 Stationary arm & 50.2 $\pm$ 0.4 & 18.1  $\pm$ 1.1&309.6 $\pm$  3.1 \\
\textbf{Mobile arm} & \textbf{82.6 $\pm$ 5.3} & \textbf{28.8 $\pm$ 4.0} & \textbf{167.1 $\pm$ 26.7} \\

\bottomrule
\end{tabular}
\end{table}
\begin{figure}[!bp]
\includegraphics[width=0.24\textwidth]{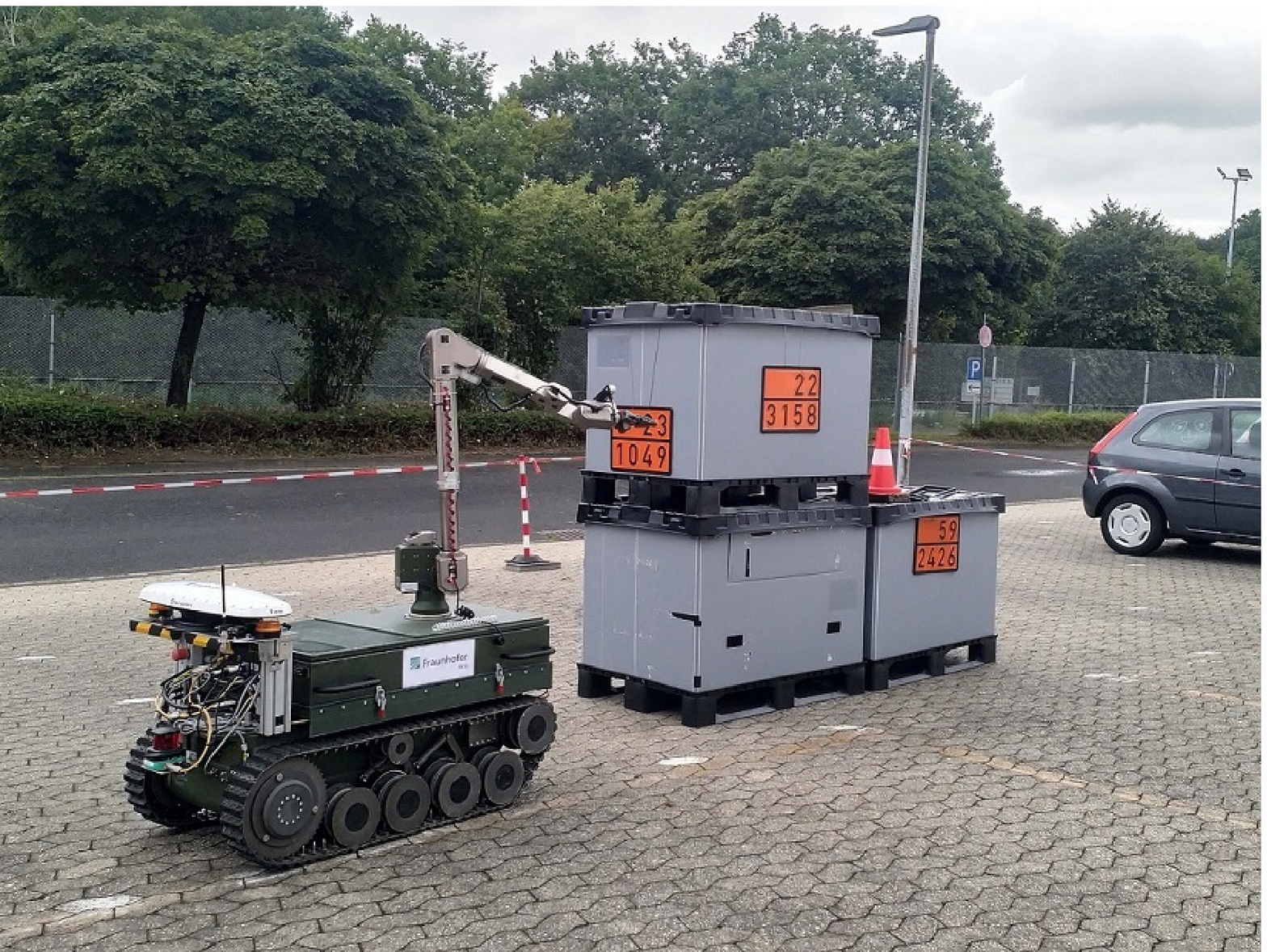} \includegraphics[width=0.24\textwidth]{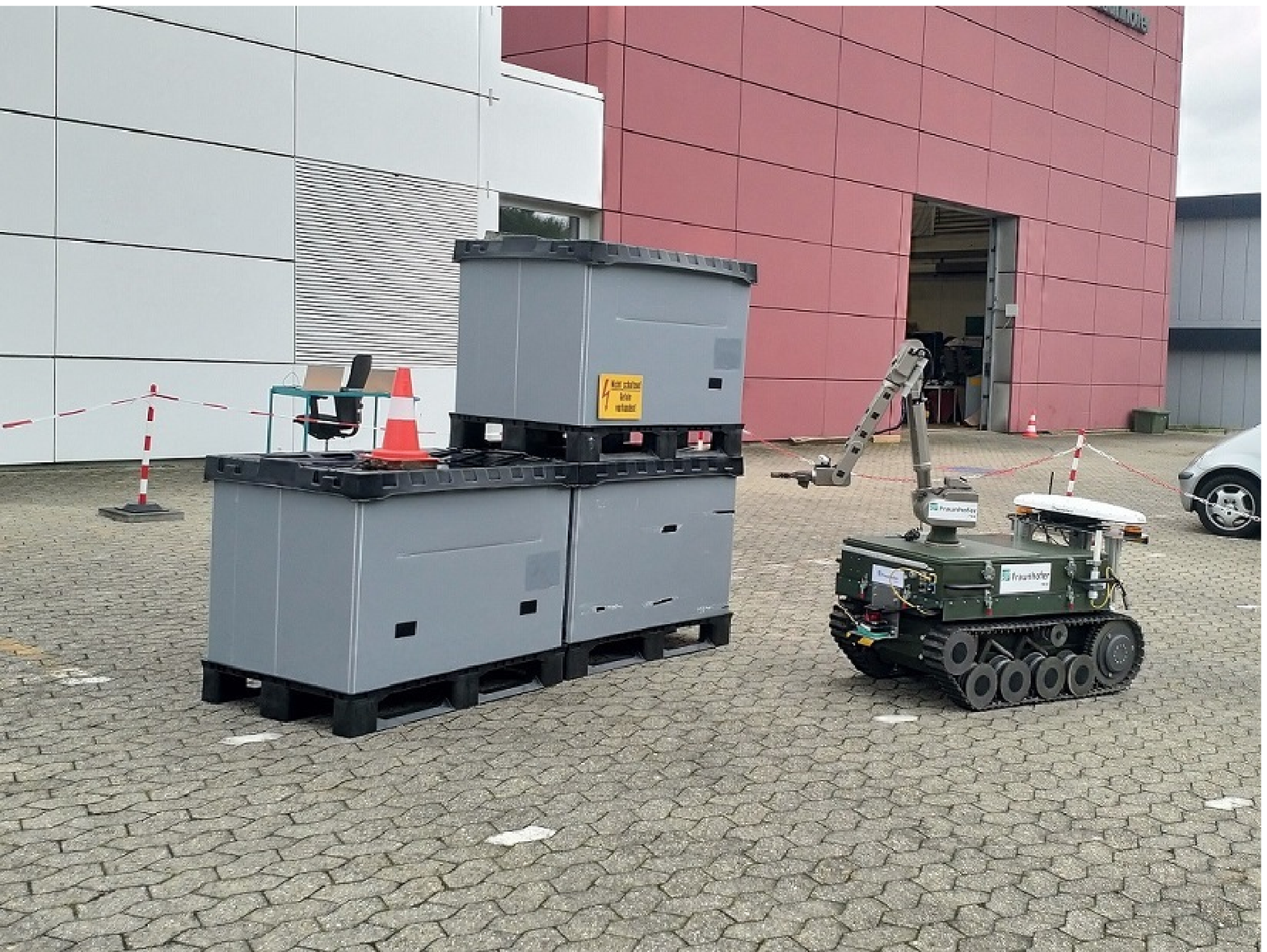}
\caption{Exploration and inspection of ROI by the real robot.}
	\label{fig:real_tests} 
	\vspace{-2ex}
\end{figure}
\subsection{Real World Outdoor Experiments}
\noindent In additional experiments we applied our approach on a real ground robot.
The robot used is a Telerob tEODor equipped with a Telerob teleMAX manipulator arm with 7 degrees of freedom.
For volumetric mapping, a RealSense D455 camera was mounted near the EEF.
The WG-NBVP was run on the onboard computer which is equipped with an Intel i7-7700T\,CPU owning 4 cores running at 2.9\,GHz and 16\,GB of RAM.
We conducted the experiments in an outdoor setting at the premises of Fraunhofer FKIE.
The exploration area was about 9\,m $\times$ 6\,m in size containing plastic pallet boxes at its center. 
To create the ROI, we generated simulated contamination measurements $m_i$ in such a way that the intensity map $\mathcal{I}$ overlaps with the plastic boxes. 

\begin{figure}[!tp]
\includegraphics[width=0.25\textwidth]{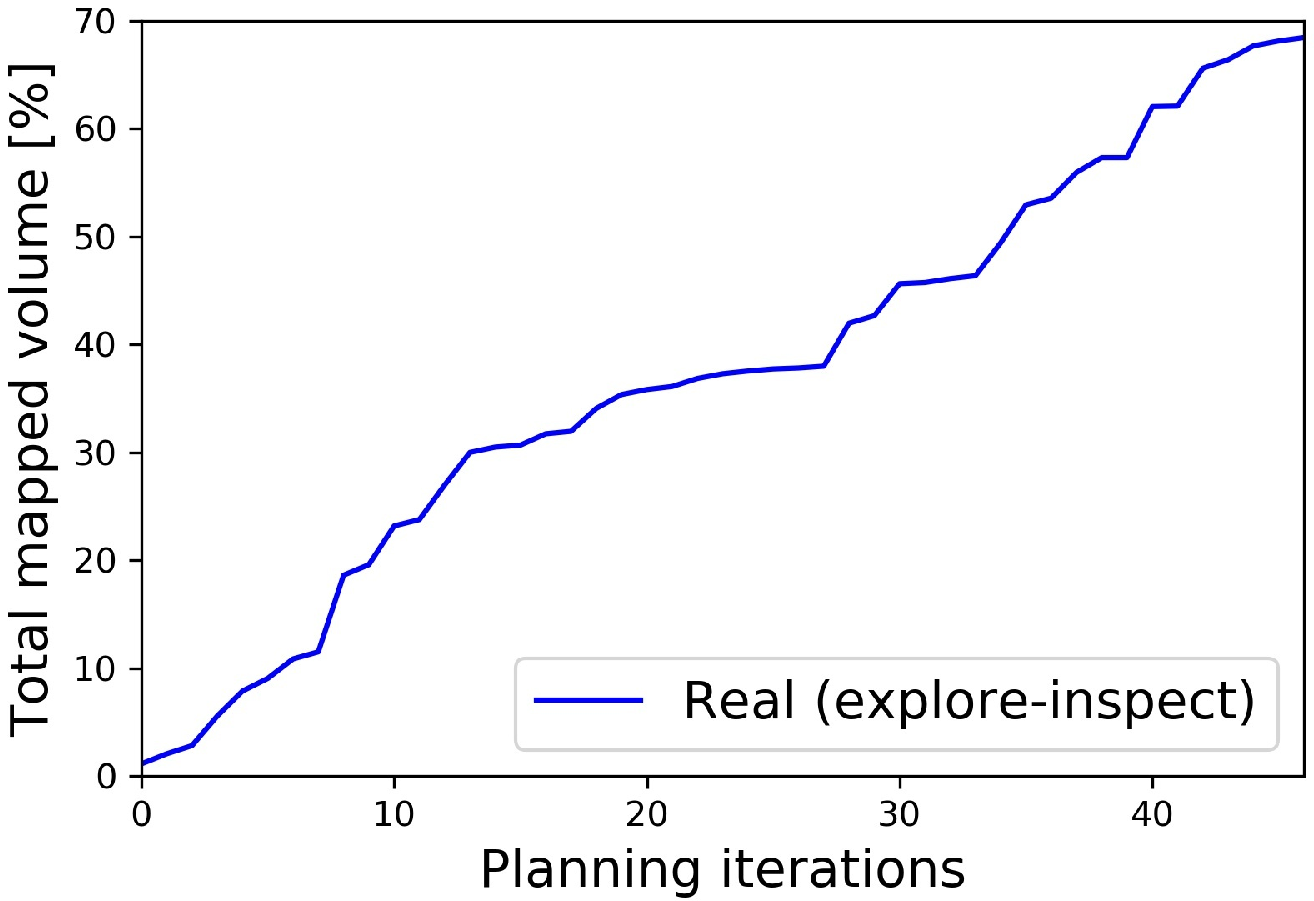}\includegraphics[width=0.25\textwidth]{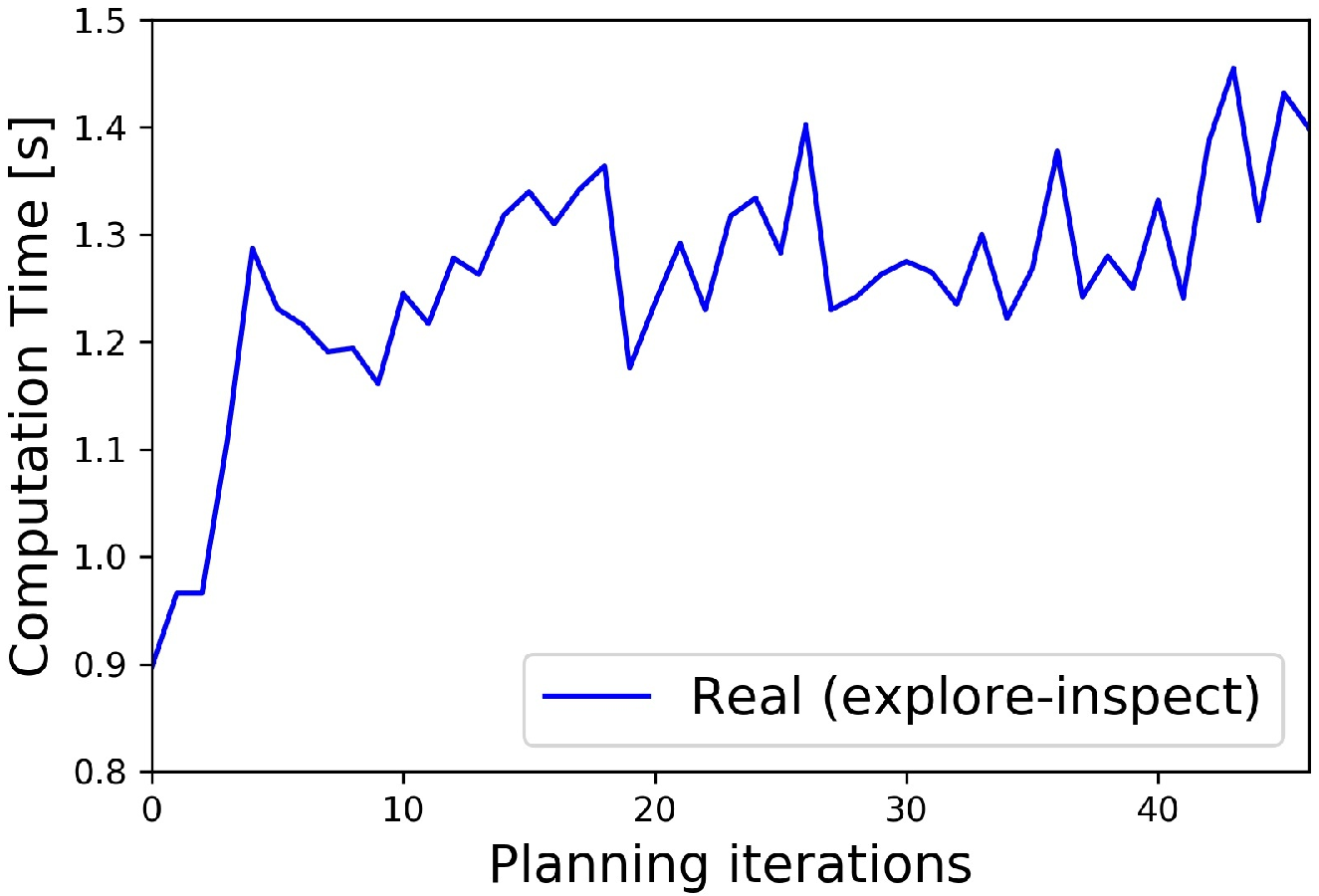} 
	\caption{Results obtained by the real mobile manipulator robot for the task of exploration-inspection during the outdoor tests.}
	\label{fig:real_plots}
	\vspace{-2ex}
\end{figure}

Fig.~\ref{fig:real_tests} shows the mobile manipulator robot performing the task of exploration-inspection. 
It can be seen that the camera on the arm is oriented towards the ROI to perform inspection for the end-user. 
In about 20 minutes, the robot explores the unknown environment and inspects the objects where the contamination values are high.
The results collected during the real experiment are shown in Fig.~\ref{fig:real_plots}.
Within 50 planning iterations, about 70\% of the environment is explored (left) and the computation time (right) is even faster than during the simulations. Each planning request takes less than 1.5 seconds despite increasing the value of $N_{max}$ to 600 nodes. 
On an average about 83 nodes were expanded per iteration with the largest tree size of 150 nodes.
Fig.~\ref{fig:teaser} shows the generated OctoMap~\cite{hornung2013octomap} that can also be visualised online by the end-user as the exploration progresses along with the sensor measurements of the ROI.

\section{CONCLUSIONS}

\noindent In this work, we proposed a novel online NBV planner for a mobile manipulator robot which can execute both exploration-inspection tasks using prior known $m_i$ values and generic exploration tasks. 
This is one of the first methods that addresses the proposed exploration-inspection problem using a mobile manipulator robot and demonstrates the practicability of the approach by transferring it to a real robot. 
We use a weighted-sum-based information gain function to solve exploration problems with multiple objectives. 
Different weights can be assigned to prioritize the needs of the end-user and guide the exploration.  
We validate the versatility of our proposed WG-NBVP for both the exploration tasks through simulation experiments in different scenarios. 
WG-NBVP shows improvements in terms of total volume mapped and lower computation time for planning. The real world experiments with the mobile manipulator robot demonstrate the ability of WG-NBVP to perform exploration and inspection of the ROI for the end-user in an outdoor setting.

\end{document}